\documentclass[preprint,times]{elsarticle}

\usepackage{amssymb}
\usepackage[figuresright]{rotating}
\usepackage[bookmarks=false]{hyperref}
    \hypersetup{colorlinks,
      linkcolor=blue,
      citecolor=blue,
      urlcolor=blue}
\usepackage{graphicx}
\usepackage{subcaption}
\usepackage{algorithm}
\usepackage{algpseudocode}
\usepackage{amsmath,amssymb,amsfonts}
\usepackage{booktabs}

\begin{document}
\begin{frontmatter}

\title{Intelligent Truck Matching in Full Truckload Shipments using Ping2Hex approach}

\author[a]{Srinivas Kumar Ramdas}
\author[b]{Jose Mathew}
\author[b]{Ankit Singh Chauhan}
\author[b]{Dinesh Rajkumar}
\author[b]{Aravind Manoj}
\author[b]{Mohit Goel\corref{cor1}}

\address[a]{Project44 Gmbh, Berlin, Germany}
\address[b]{Project44 Inc, Bengaluru, India}

\cortext[cor1]{Corresponding author. Email: mohit.goel@project44.com}

\begin{abstract}
Accurate truck-to-shipment matching using GPS data is foundational for full truckload supply chain visibility, enabling real-time tracking and accurate estimated time of arrival (ETA) predictions. However, missing or corrupted vehicle identifiers prevent traditional matching approaches, leaving shipments without visibility. This paper presents Intelligent Truck Matching (ITM) 2.0, a machine learning system that addresses this critical gap by formulating matching as a probabilistic ranking problem. Our approach leverages Uber H3 hexagonal spatial indexing to discretize GPS pings into route similarity features, combined with temporal information, then applies LightGBM gradient boosting with threshold-based post-processing.

Through rigorous evaluation including offline model selection (SVM, XGBoost, LightGBM), comprehensive ablation studies, and production shadow testing, we demonstrate substantial gains over rule-based baselines. ITM 2.0 achieves 26 percentage point precision improvement in North America and 14 points in Europe, while doubling coverage. Deployed in production at Project44 handling full truckload shipments, the system demonstrates robustness to geocoding errors up to 1 km, multiple candidate trucks, and sparse pings. These improvements directly translate to enhanced operational capabilities: hundreds more full truckload shipments gain real-time tracking daily, enabling accurate ETA predictions, proactive exception management, and improved customer service in the trucking industry.
\end{abstract}


\end{frontmatter}

\noindent\textit{Accepted at the International Conference on Industry Sciences and Computer Sciences Innovation (iSCSi) 2026. To appear in Procedia Computer Science (Elsevier).}

\bigskip

\noindent\textit{Author emails:} srinivaskumar.ramdas@project44.com (S.K. Ramdas, first author), jose.mathew@project44.com (J. Mathew), ankitsingh.chauhan@project44.com (A.S. Chauhan), dinesh.rajkumar@project44.com (D. Rajkumar), aravind.manoj@project44.com (A. Manoj), mohit.goel@project44.com (M. Goel)

\section{Introduction}
\label{sec:introduction}

Real-time visibility of truck location and shipment status is critical for efficient full truckload freight management, enabling accurate estimated time of arrival (ETA) predictions, proactive exception management, and improved customer service. The adoption of electronic logging devices (ELDs) and GPS-based tracking has provided location visibility \cite{ELD}. However, missing or incorrect truck identifiers—such as vehicle IDs or ELD device IDs—severely hinder shipment visibility. These gaps, caused by device errors, data loss, or manual entry inconsistencies, prevent associating GPS pings with shipments, leaving customers without tracking information and carriers unable to provide reliable ETAs \cite{tsolaki,IFTO}.

In large-scale full truckload operations monitoring hundreds of daily shipments, traditional matching methods fail under these conditions \cite{barnhart1999,chen2024}, preventing accurate ETA calculation, delaying exception alerts, and directly impacting service levels.

Recent work in machine learning, spatial indexing, and trajectory analysis has shown promise in tackling spatial matching problems in complex, data-rich environments. Frameworks like Uber's H3 hexagonal spatial index \cite{uber_h3} and advanced tree-based models such as LightGBM \cite{lgbm,LGBM_THEO} have enabled better geospatial feature extraction and robust predictive modeling for logistics tasks. However, existing solutions rarely address the truck-shipment matching challenge when explicit identifiers are unavailable and geocoding noise is not tackled either.

\begin{figure}[!htbp]
    \centering
    \includegraphics[width=0.45\textwidth]{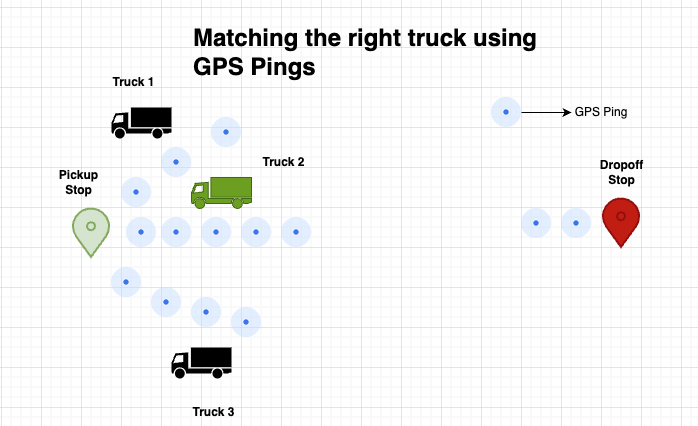}
    \caption{Choosing the right truck from a set departing from the same pickup stop using GPS pings.}
    \label{fig:itm-truck_matching}
\end{figure}

We formulate this as a task of selecting the correct truck from among candidates departing from the pickup stop, based on the likelihood that it will reach the intended destination within the expected timeframe, as illustrated in Fig.~\ref{fig:itm-truck_matching}.

In this paper, we present a novel \emph{Intelligent Truck Matching} (ITM) system that formulates the matching task as a probabilistic ranking problem over candidate pingsets. Our pipeline discretizes GPS trajectories using hexagonal cells, deriving route similarity and temporal features to predict the most likely truck for each shipment. The proposed system is deployed at scale in production, demonstrating significant improvements in precision and coverage over prior methods and exhibiting robustness to data gaps and location inaccuracies.

\subsection{Main Contributions}

Our main contributions are:
\begin{itemize}
    \item A formal problem formulation and end-to-end pipeline for GPS-based truck-shipment matching which is unique in supply chain literature
    \item The ITM architecture ML pipeline which improves the truck matching coverage and accuracy over incumbent rule based engine
    \item A scheme for dataset construction, using hexagonal grid-based spatial feature extraction method (ping2hex), improving route similarity modeling \cite{uber_h3}
\end{itemize}

\section{Related Work}

\subsection{Machine Learning in Freight Logistics}

Machine learning has been applied to freight challenges including shipment visibility and ETA prediction \cite{IFTO}, with appointment times serving as critical temporal anchors \cite{appmnt}. Gradient boosting methods, particularly LightGBM, have shown strong results in supply chain forecasting by combining temporal and geospatial features \cite{lgbm,LGBM_THEO}. However, prior work predominantly assumes that truck identifiers are available and correct—focusing on optimizing routing, scheduling, or demand forecasting rather than the upstream problem of identifying which truck is carrying which shipment. This leaves a critical gap: when identifiers are missing or corrupted, downstream visibility and ETA systems have no input to work with. Our work addresses this gap directly by treating truck matching as a classification problem over GPS trajectories.

\subsection{Spatial Indexing for Trajectory Analysis}

Geospatial discretization is essential for comparing trajectories at scale. Uber's H3 hexagonal indexing \cite{uber_h3} offers advantages over square grids and geohashes: uniform cell areas, consistent neighbor relationships, and hierarchical multi-resolution support. While H3 has been adopted for ride-sharing demand prediction and urban analytics, its application to freight trajectory matching is unexplored in the literature. We leverage H3's properties specifically for route similarity computation—a novel use case that exploits the hierarchical structure to balance geocoding noise tolerance with spatial precision.

\subsection{Truck-Shipment Matching}

Truck-to-shipment matching has traditionally been formulated as a combinatorial optimization problem, solved via heuristics or integer programming \cite{barnhart1999,tang2024}. Recent work has introduced ML-guided approaches, such as Q-learning for dock matching \cite{li2024} and online matching for less-than-truckload logistics \cite{chen2024}. These methods assume accurate inputs—known truck identities, service constraints, and cost functions. In contrast, the matching problem we address involves missing or noisy identifiers and geospatial uncertainties, requiring a fundamentally different formulation. Rather than optimizing assignment costs, we predict the probability that a candidate GPS trajectory belongs to a given shipment, framing matching as a ranking problem under uncertainty.

\section{Problem Formulation}
\label{sec:prob_form}

A truck's pingset is the set of GPS pings received from the ELD device of a single truck over a time period defined by the appointment windows of the shipment.

We define the task of \emph{truck matching} as a probabilistic ranking problem: for a given shipment $S$, we select the truck pingset $P^*$ from a set of candidates $\{P_1, P_2, \dots, P_N\}$ that maximizes the probability of being the correct match, as estimated by our model. In other words, we choose the pingset for which this probability is highest.

\begin{equation}
P^* = \underset{P_i}{\arg\max}\;\; \Pr(P_i \mid S)
\label{eq:truck-matching}
\end{equation}

\noindent
where:
\begin{itemize}
    \item $P^*$: the selected truck pingset that is the best match for shipment $S$
    \item $P_i$: the $i$-th candidate pingset from the set $\{P_1, P_2, \dots, P_N\}$
    \item $S$: the shipment to be matched
    \item $\Pr(P_i \mid S)$: the probability, as predicted by our model, that pingset $P_i$ correctly matches shipment $S$
\end{itemize}

\begin{figure}[!htbp]
    \centering
    \begin{subfigure}[b]{0.32\textwidth}
        \centering
        \includegraphics[width=\textwidth]{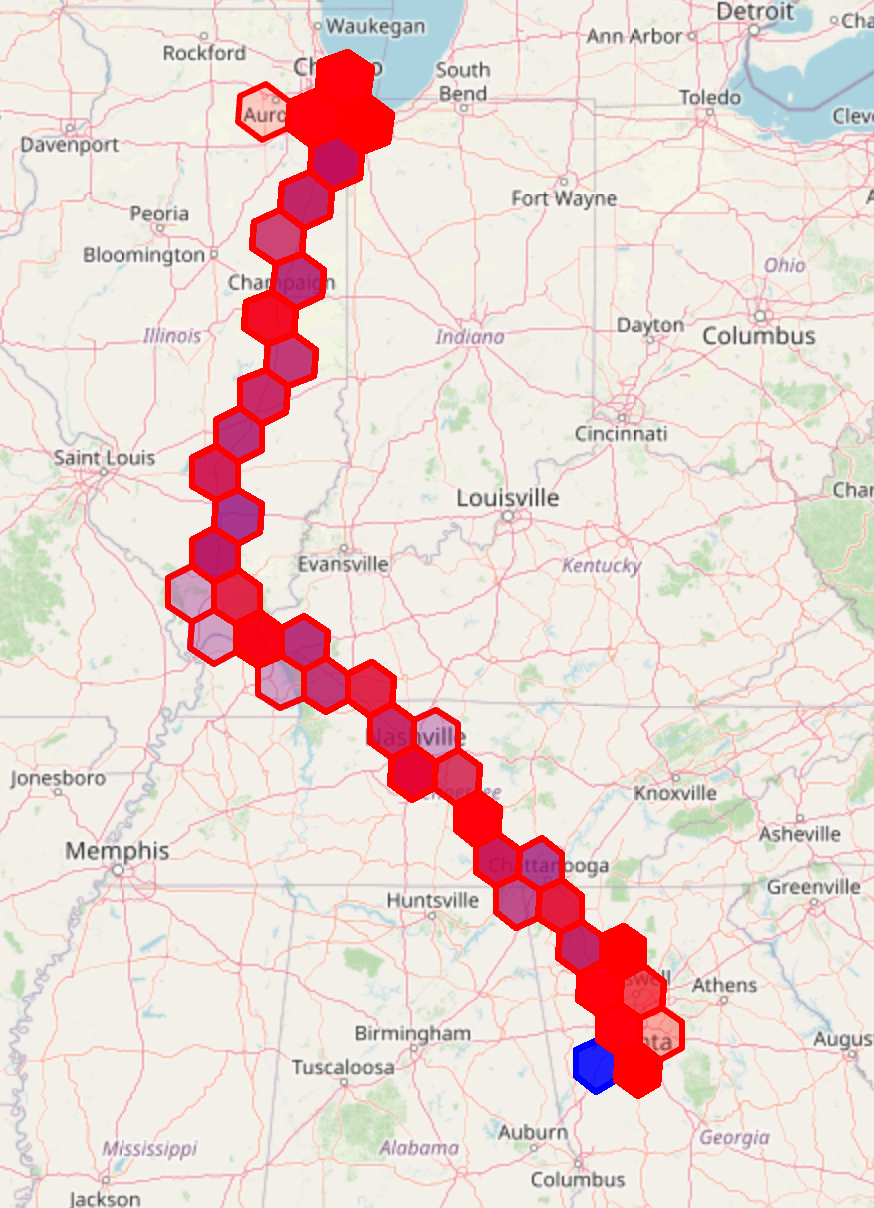}
        \caption{Good match: many hexcells overlap.}
        \label{fig:good-match}
    \end{subfigure}%
    \hfill
    \begin{subfigure}[b]{0.32\textwidth}
        \centering
        \includegraphics[width=\textwidth]{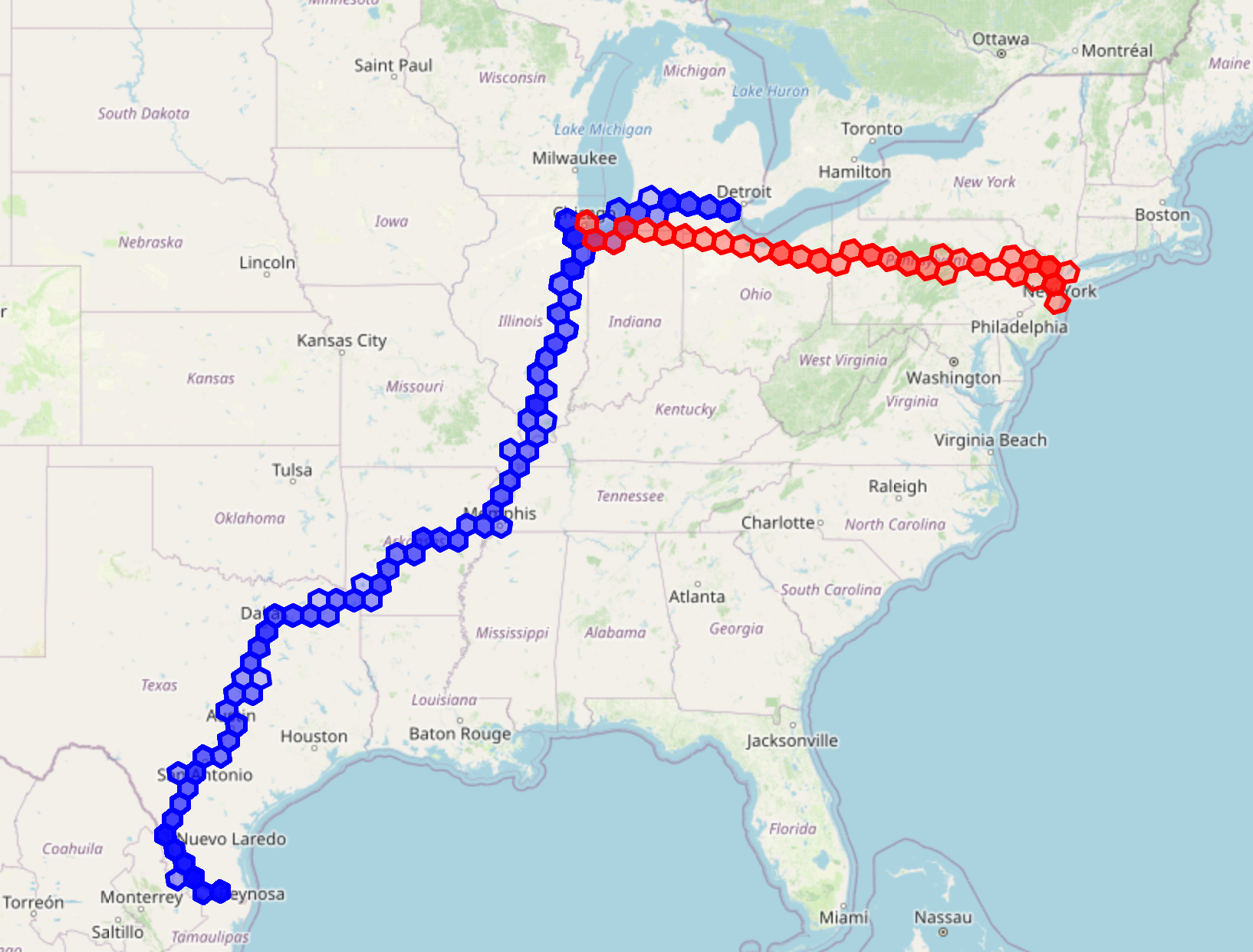}
        \caption{Poor match: few hexcells overlap.}
        \label{fig:bad-match}
    \end{subfigure}
    \caption{Pingsets mapped to hexcells. Red: historical lane; blue: current truck. Overlap count determines match quality.}
    \label{fig:comparison}
\end{figure}

Fig.~\ref{fig:comparison} illustrates good and poor matches: truck pings and historical lane pings are both mapped to hexcells, and the degree of hexcell overlap captures route similarity as a geospatial feature.

Accurate truck matching depends on the number of GPS pings available, the accuracy of these pings (map providers are prone to geocoding errors), and the truck's adherence to appointment schedules.

\section{ITM Assignment Pipeline}

\begin{figure}[!htbp]
    \centering
    \includegraphics[width=0.5\textwidth]{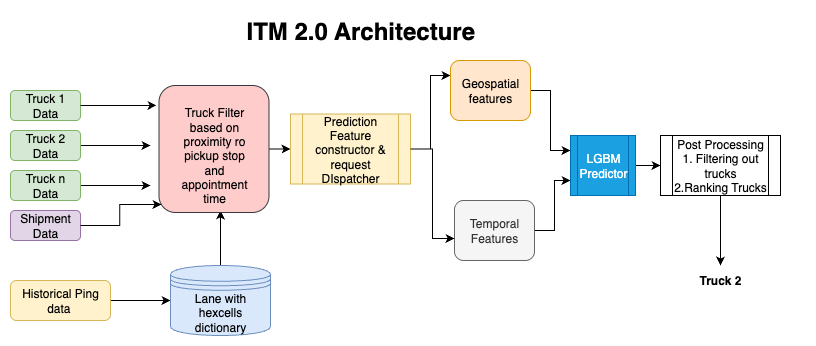}
    \caption{ITM 2.0 architecture: from candidate truck filtering to LGBM prediction and post-processing.}
    \label{fig:itm-architecture}
\end{figure}

The ITM pipeline (Fig.~\ref{fig:itm-architecture}) maps lane pings to hexcells, discretizing geographic information, and uses appointment time information to match trucks to shipments. A lane comprises a unique pair of origin and destination city locations represented as hexcells. We employ LightGBM for probability prediction, as tree-based models have shown strong performance in similar tasks \cite{tsolaki}.

For a given shipment, we extract appointment timestamps, ping data, and temporal features. Carrier details identify candidate trucks, which are filtered by proximity to the pickup location and appointment time adherence. Using shipment pickup and drop-off information, we extract historical lane data and construct geospatial and temporal features (Table~\ref{tab:feature_composition}).

\begin{table}[h]
\caption{Feature composition for ITM 2.0 model}
\label{tab:feature_composition}
\begin{tabular*}{\hsize}{@{\extracolsep{\fill}}lll@{}}
\toprule
\textbf{Feature Name} & \textbf{Category} & \textbf{Description} \\
\midrule
Ping time difference to pickup & Temporal & Time elapsed from scheduled pickup \\
& & appointment to current ping timestamp \\
\midrule
Ping distance to destination & Geospatial & Euclidean distance (km) from current \\
& & ping location to delivery destination \\
\midrule
Number of overlapping hexcells & Hexcell-based & Count of H3 hexcells traversed by \\
& Geospatial & both current truck and historical lane \\
\midrule
Number of pings within & Hexcell-based & Total GPS pings within overlapping \\
overlapped hexcells & Geospatial & hexcells (confidence measure) \\
\bottomrule
\end{tabular*}
\end{table}

For each candidate truck, features are computed and fed to the LGBM predictor, which outputs match probability. Predictions occur at periodic intervals as new pings arrive, with confidence increasing as trajectory information accumulates. All predictions feed into a post-processing layer that ranks trucks exceeding a probability threshold; if none pass, we wait for more pings.

\subsection{Historical Lane ping Data Construction}

For constructing the historical lane ping data, we map each ping to its corresponding hexcell using a ping2hex approach. We use H3 resolution 4 for origin and destination cities (26 km edge length) and resolution 6 for individual pings (3.72 km edge length) \cite{h3_res}. This choice ensures that pickup and drop-off locations encompass average town/city sizes while allowing precise trajectory tracking along the lane.

\begin{figure}[!htbp]
    \centering
    \begin{subfigure}[b]{0.32\textwidth}
        \centering
        \includegraphics[width=\textwidth]{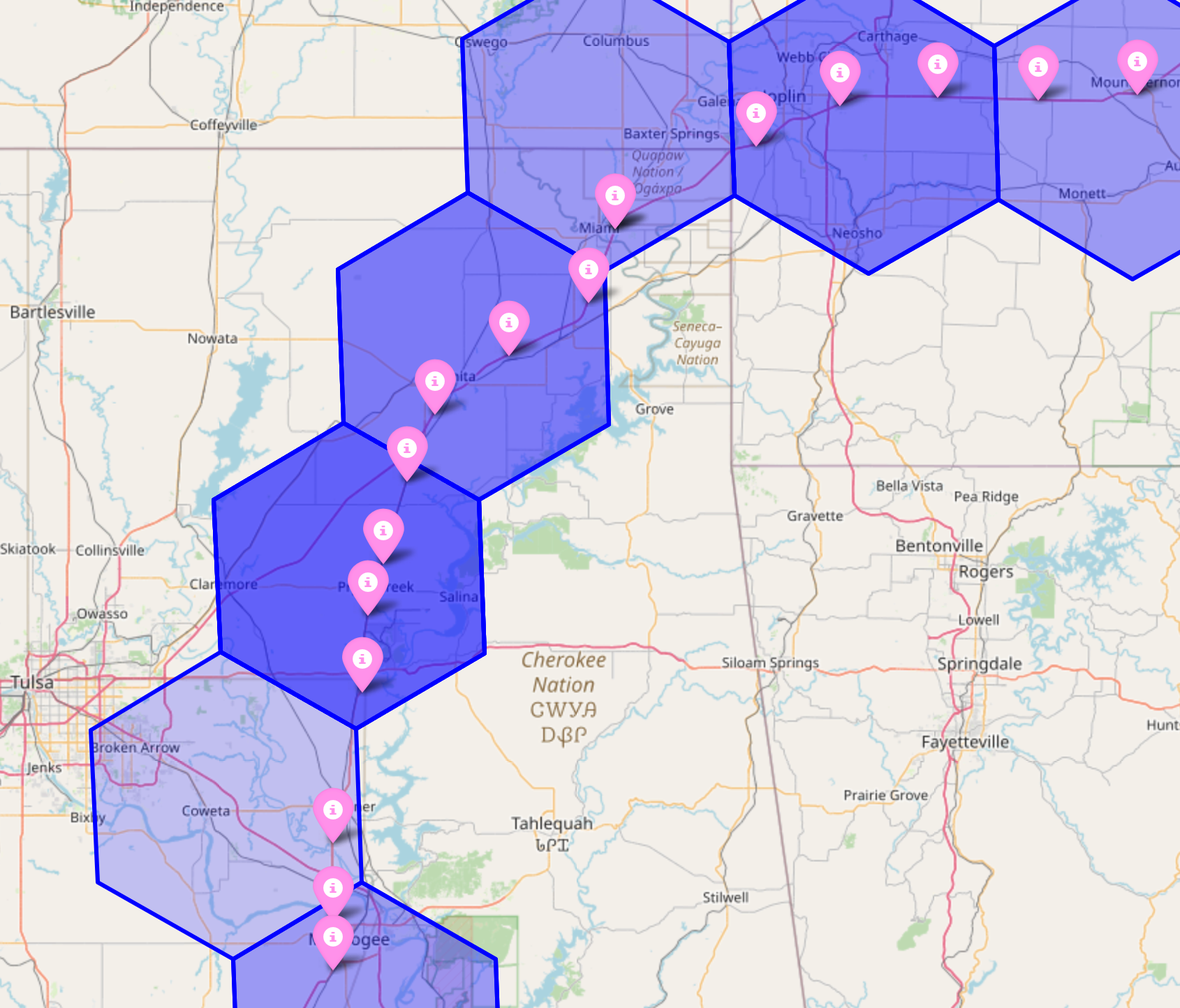}
        \caption{Pings mapped to hexcells.}
        \label{fig:ping2hex}
    \end{subfigure}%
    \hfill
    \begin{subfigure}[b]{0.32\textwidth}
        \centering
        \includegraphics[width=\textwidth]{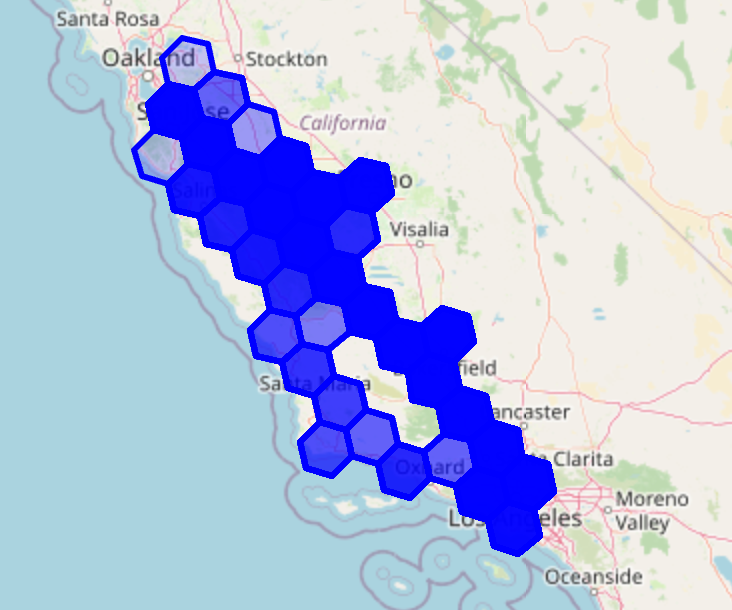}
        \caption{SJO to LA lane with multiple routes.}
        \label{fig:sjo-la}
    \end{subfigure}
    \caption{Mapping pings to hexcells and constructing the lane pings dataset.}
    \label{fig:lane_mappings}
\end{figure}

Each entry maps a lane code (hexcell hash of origin-destination) to hexcells traversed along that route. Our dataset contains approximately 2M unique lanes. Fig.~\ref{fig:sjo-la} shows a San Jose to Los Angeles lane, and Fig.~\ref{fig:ping2hex} illustrates ping-to-hex mapping. Since multiple paths exist for the same city pair, we model route similarity using hexcell overlap between the current truck and historical lane. Multiple pings within a hexcell increase confidence that the truck traversed it, so we use total pings within overlapping hexcells as a feature.

\section{Post Processing Pipeline}

The model may assign high probabilities to multiple trucks. To address this, a post-processing algorithm (Algorithm~\ref{alg:postprocessing}) ranks candidates and categorizes them into HIGH, MEDIUM, and LOW confidence classes. Trucks are only assigned if they exceed a minimum probability threshold. Due to ping sparsity, the post-processing layer can select lower-confidence candidates when information is limited, communicating confidence levels to downstream users.

\begin{algorithm}[H]
\small
\caption{Postprocessing of Trucks Based on Probability Scores}
\label{alg:postprocessing}
\begin{algorithmic}[1]
\Require List of $n$ trucks with scores: $T = \{(t_i, p_i)\}$; thresholds: $\tau_{\text{min}}$, $\tau_{\text{high}}$, $\tau_{\text{medium}}$
\Ensure Confidence labels (\texttt{LOW}/\texttt{MEDIUM}/\texttt{HIGH}) and truck assignment
\State $T' \leftarrow \{(t_i, p_i) \in T \mid p_i \geq \tau_{\text{min}}\}$
\If{$T' = \emptyset$} \textbf{return} \texttt{LOW}, no truck assigned
\EndIf
\State Sort $T'$ by $p_i$ descending
\ForAll{$(t_i, p_i) \in T'$}
    \If{$p_i \geq \tau_{\text{high}}$} label $\gets$ \texttt{HIGH}
    \ElsIf{$p_i \geq \tau_{\text{medium}}$} label $\gets$ \texttt{MEDIUM}
    \Else{} label $\gets$ \texttt{LOW}
    \EndIf
\EndFor
\State $t^* = \arg\max_{t_i} p_i$
\State \textbf{return} confidence labels and assigned truck $t^*$
\end{algorithmic}
\end{algorithm}

\section{Training and Deployment}
\label{sec:training}

Training was conducted on Google Vertex AI \cite{google_vertexai} (N2-series, 128 GB RAM, 16 vCPUs) using the LightGBM Python package. We optimize binary log loss \cite{bin_loss}, which measures the negative log-likelihood of true labels given predicted probabilities and is well-suited for our task since the model outputs calibrated match probabilities:

\begin{equation}
\mathcal{L}_{\text{binary\_logloss}} = -\frac{1}{N} \sum_{i=1}^{N} \left[ y_i \log(p_i) + (1 - y_i) \log(1 - p_i) \right]
\end{equation}

\begin{equation}
p_i = \frac{1}{1 + \exp(-f(x_i))}
\end{equation}

\noindent where $y_i \in \{0,1\}$ is the true label, $p_i$ is the predicted probability, and $f(x_i)$ is the model's raw output. An inherent precision-recall tradeoff exists \cite{prerec,prerec2}; we prioritize precision over recall (see Section~\ref{sec:metrics}).

To validate performance before full rollout, we conducted shadow testing \cite{shadow} by duplicating production requests to both systems. During shadow testing, we tuned the minimum distance threshold from origin, minimum pings required for prediction, and probability cutoffs.

Lane mappings are kept current via a daily job that adds new lanes and enriches existing ones with new hexcells, using a rolling one-year window. We cache lane mappings in Redis \cite{redis} for fast access. Inference averages 300 milliseconds per request across multiple 4-core servers.

\section{Experiments}

\subsection{Evaluation Strategy and Baseline Methods}

Our evaluation follows a two-stage approach: first, offline evaluations comparing multiple ML models to select the best algorithm; second, production shadow testing of the selected model against the incumbent ITM 1.0 baseline to validate real-world performance.

\subsubsection{Production Baseline: ITM 1.0}
\label{subsec:itm_v1}

ITM 1.0 is the incumbent rule-based truck matching system deployed in production prior to this work, serving as our primary baseline. It employs a rigid 500-meter proximity threshold from the pickup point and temporal windows around scheduled appointment times, but does not incorporate transit trajectory information once a truck departs nor leverage historical lane data.

ITM 1.0 struggles in two critical scenarios: (1) when multiple candidate trucks depart from the same location but diverge en route to different destinations, and (2) in the presence of geocoding errors. The stringent 500-meter threshold makes it highly sensitive to geocoding inaccuracies \cite{map_error}, which are prevalent with typical map providers and warehouse environments that often span areas beyond one kilometer, as illustrated in Fig.~\ref{fig:geo_error}. Despite these limitations, ITM 1.0 represents a strong baseline refined over multiple production iterations.

\subsubsection{Machine Learning Baselines: Offline Model Selection}
\label{subsec:model_selection}

To select the most suitable machine learning algorithm for the truck matching task, we conducted comprehensive offline evaluations comparing three diverse model families on identical feature sets and training data:

\begin{itemize}
    \item \textbf{Support Vector Machine (SVM) Regressor} \cite{svm}: A kernel-based method capable of learning non-linear decision boundaries
    \item \textbf{XGBoost} \cite{xgboost}: A popular gradient boosting framework with level-wise tree growth
    \item \textbf{LightGBM} \cite{LGBM_THEO}: A gradient boosting framework with leaf-wise tree growth and histogram-based learning
\end{itemize}

Table~\ref{tab:model_comparison} presents the comparative results. SVM Regressor performed poorly due to high dimensionality, class imbalance, and lack of calibrated probabilities. LightGBM consistently outperformed XGBoost by ~10 percentage points in both metrics across regions, with 3× faster training and lower memory consumption. LightGBM's leaf-wise tree growth and histogram-based learning handle the high-dimensional hexcell feature space and class imbalance (150K positive vs 350K negative) effectively, producing well-calibrated probabilities for Algorithm~\ref{alg:postprocessing}. We selected LightGBM as the base model for ITM 2.0.

\begin{table}[h]
\caption{Offline model comparison on validation set across three ML approaches}
\label{tab:model_comparison}
\begin{tabular*}{\hsize}{@{\extracolsep{\fill}}llll@{}}
\toprule
\textbf{Model} & \textbf{Region} & \textbf{Coverage} & \textbf{Precision} \\
\midrule
SVM Regressor & NA & 5.2\% & 18\% \\
XGBoost & NA & 8.0\% & 45\% \\
\textbf{LightGBM (Selected)} & NA & \textbf{9.0\%} & \textbf{55\%} \\
\midrule
SVM Regressor & EU & 12\% & 12\% \\
XGBoost & EU & 22\% & 22\% \\
\textbf{LightGBM (Selected)} & EU & \textbf{27\%} & \textbf{32\%} \\
\bottomrule
\end{tabular*}
\end{table}

\subsection{Dataset}

The dataset combines actual correct truck-shipment assignments (positive samples) and simulated incorrect candidates (negative samples).

\begin{figure}[!htbp]
    \centering
    \begin{subfigure}[b]{0.30\textwidth}
        \centering
        \includegraphics[width=\textwidth]{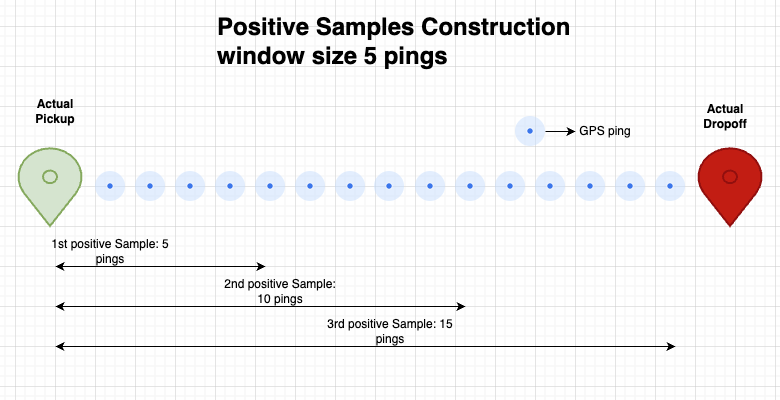}
        \caption{Positive: windowed snapshots.}
        \label{fig:pos_samples}
    \end{subfigure}%
    \hfill
    \begin{subfigure}[b]{0.30\textwidth}
        \centering
        \includegraphics[width=\textwidth]{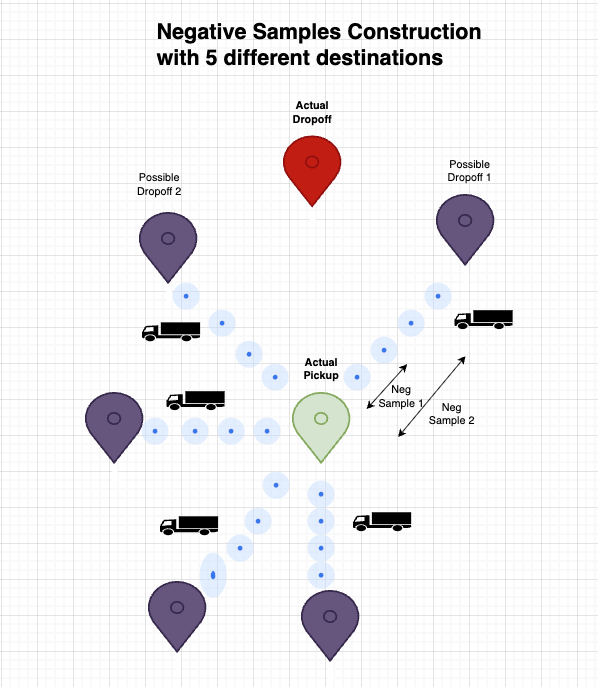}
        \caption{Negative: different destinations.}
        \label{fig:neg_samples}
    \end{subfigure}
    \caption{Dataset sample construction using snap-shotting from actual and alternative truck journeys.}
    \label{fig:dataset_samples}
\end{figure}

For positive samples, we removed known truck-shipment associations and took windowed ping snapshots at intervals (20, 25, 30 pings, etc.). As shown in Fig.~\ref{fig:pos_samples}, this trains the model to predict early in the journey. Pings contain geospatial and temporal features (Table~\ref{tab:ping_features}).

For negative samples, we used trucks approaching the origin but traveling to different destinations. This captures trucks that initially overlap in trajectory but later diverge (Fig.~\ref{fig:neg_samples}). We included at least 5 alternative destinations per origin-destination pair for robustness.

\begin{table}[h]
\caption{Ping feature composition}
\label{tab:ping_features}
\begin{tabular*}{\hsize}{@{\extracolsep{\fill}}ll@{}}
\toprule
\textbf{Category} & \textbf{Description} \\
\midrule
Location & Country, state, city \\
Geographic features & Latitude and longitude of ping \\
Temporal features & UTC timestamp of ping, offset of local timezone \\
Label & Truck ID the ping belongs to \\
\bottomrule
\end{tabular*}
\end{table}

The dataset spans a 3-month period, yielding 150K positive samples and 350K negative samples. We intentionally skewed toward more negative samples to reflect the higher cost of false positives in production.

\subsubsection{Lane Dataset}

We used one year of ping data to construct the lane mappings, resulting in 760K unique lanes from 57M pings across 5.6M shipments. The data structure is a dictionary where each lane code (unique origin-destination pair) maps to a list of hexcells traversed along that route.

\subsection{Results}

We present quantitative comparisons to baselines using performance metrics and qualitative analysis of different scenarios.

\subsection{Metrics}
\label{sec:metrics}

We evaluate using \emph{coverage} ($N_{\text{assigned}} / N_{\text{eligible}}$) and \emph{precision} ($N_{\text{correct}} / N_{\text{assigned}}$), where a correct assignment is one where the truck visits both origin and destination within 2 hours of appointment times. We prioritize precision because assigning the wrong truck is more costly than making no assignment.

\subsubsection{Quantitative Results: Production Shadow Testing}

We deployed ITM 2.0 in production shadow mode \cite{shadow}, duplicating actual requests to both systems simultaneously for direct comparison without impacting live operations. Table~\ref{tab:comp_baselines_aggregate} presents aggregate results across North America (NA) and European Union (EU), reported separately due to different operational characteristics. Table~\ref{tab:comp_baselines_tenantwise} breaks down performance by individual high-volume customer tenants with varying haul distances.

\begin{table}[h]
\caption{Production shadow testing results: ITM 2.0 vs ITM 1.0 (Aggregate across NA and EU regions)}
\label{tab:comp_baselines_aggregate}
\begin{tabular*}{\hsize}{@{\extracolsep{\fill}}llll@{}}
\toprule
\textbf{Region} & \textbf{Method} & \textbf{Coverage} & \textbf{Precision} \\
\midrule
NA & ITM 1.0 & 4.5\% & 29\% \\
NA & ITM 2.0 & 9.0\% & \textbf{55\%} \\
EU & ITM 1.0 & 2.5\% & 18\% \\
EU & ITM 2.0 & 27\% & \textbf{32\%} \\
\bottomrule
\end{tabular*}
\end{table}

\begin{table}[h]
\caption{Production shadow testing results: ITM 2.0 vs ITM 1.0 (Tenant-wise breakdown)}
\label{tab:comp_baselines_tenantwise}
\begin{tabular*}{\hsize}{@{\extracolsep{\fill}}lllll@{}}
\toprule
\textbf{Region} & \textbf{Method} & \textbf{Tenant Type} & \textbf{Coverage} & \textbf{Precision}\\
\midrule
NA & ITM 1.0 & Long Haul Shipper ($\sim$500km) & 1.20\% & 19.74\% \\
NA & ITM 2.0 & Long Haul Shipper ($\sim$500km) & 4.94\% & \textbf{66.77\%} \\
NA & ITM 1.0 & Long Haul Shipper ($\sim$400km) & 1.59\% & 23.08\% \\
NA & ITM 2.0 & Long Haul Shipper ($\sim$400km) & 24.76\% & \textbf{58.13\%} \\
EU & ITM 1.0 & Long Haul ($\sim$500km) & 11.51\% & 22.70\% \\
EU & ITM 2.0 & Long Haul ($\sim$500km) & 17.71\% & \textbf{43.93\%} \\
EU & ITM 1.0 & Medium Haul ($\sim$250km) & 5.42\% & 33.72\% \\
EU & ITM 2.0 & Medium Haul ($\sim$250km) & 14.96\% & \textbf{67.78\%} \\
\bottomrule
\end{tabular*}
\end{table}

\subsubsection{Ablation Study}
\label{subsec:ablation}

We conducted an ablation study by removing key components (Table~\ref{tab:ablation_study}). Removing post-processing drops precision from 55\% to 20\% in NA and 32\% to 10\% in EU. Removing hexcell features entirely (temporal only) causes precision to fall to 7\% in NA and 6\% in EU, confirming that spatial discretization is the primary driver of matching accuracy.

\begin{table}[h]
\caption{Ablation study showing impact of removing key components}
\label{tab:ablation_study}
\begin{tabular*}{\hsize}{@{\extracolsep{\fill}}lllll@{}}
\toprule
\textbf{Model Configuration} & \textbf{Region} & \textbf{Coverage} & \textbf{Precision} & \textbf{$\Delta$ Precision} \\
\midrule
\textbf{Full ITM 2.0 (Baseline)} & NA & 9.0\% & \textbf{55\%} & -- \\
\textbf{Full ITM 2.0 (Baseline)} & EU & 27\% & \textbf{32\%} & -- \\
\midrule
Without Post-processing & NA & 13.5\% & 20\% & -35 p.p. \\
Without Post-processing & EU & 32\% & 10\% & -22 p.p. \\
\midrule
Without Hexcell Features & NA & 24\% & 7\% & -48 p.p. \\
(Temporal only) & EU & 47\% & 6\% & -26 p.p. \\
\bottomrule
\end{tabular*}
\end{table}

ITM 2.0 substantially outperforms ITM 1.0: 26 p.p. precision improvement in NA while doubling coverage, and 14 p.p. in EU with 10× coverage increase.

Model precision also varies with destination proximity percent, defined as $\text{DPP} = (d_{\text{latest\_ping} \to \text{dest}} \;/\; d_{\text{pickup} \to \text{dest}}) \times 100$, where $d_{\text{latest\_ping} \to \text{dest}}$ is the distance from the most recent ping to the destination and $d_{\text{pickup} \to \text{dest}}$ is the total pickup-to-destination distance. The median distance to destination is $\text{MDD} = \operatorname{median}( d_{i,\text{dest}} )$ across all assignments at a given threshold. Table~\ref{tab:precision_vs_distance} shows precision at different DPP thresholds. We use the 50\% DPP threshold in production to balance early matching with reliability.

\begin{table}[h]
\caption{Precision at different destination proximity thresholds}
\label{tab:precision_vs_distance}
\begin{tabular*}{\hsize}{@{\extracolsep{\fill}}lll@{}}
\toprule
\textbf{DPP} & \textbf{Precision} & \textbf{MDD (km)} \\
\midrule
$\leq 25$\% & 76\% & 2 \\
\textbf{$\leq 50$\% (Production)} & \textbf{50\%} & \textbf{30} \\
$\geq 80$\% & 35\% & 83 \\
\bottomrule
\end{tabular*}
\end{table}

\subsubsection{Short-Haul Performance}
\label{subsec:short_haul}

For short-haul shipments (distance $\leq$ 40 km), ITM 2.0 achieves approximately 20\% precision with high coverage. In production deployment, distance-based filtering excludes short-haul shipments from ITM 2.0 processing, routing them to ITM 1.0 instead.

\subsubsection{Qualitative Results}

\begin{figure}[!htbp]
    \centering
    \begin{subfigure}[b]{0.32\textwidth}
        \centering
        \includegraphics[width=\textwidth]{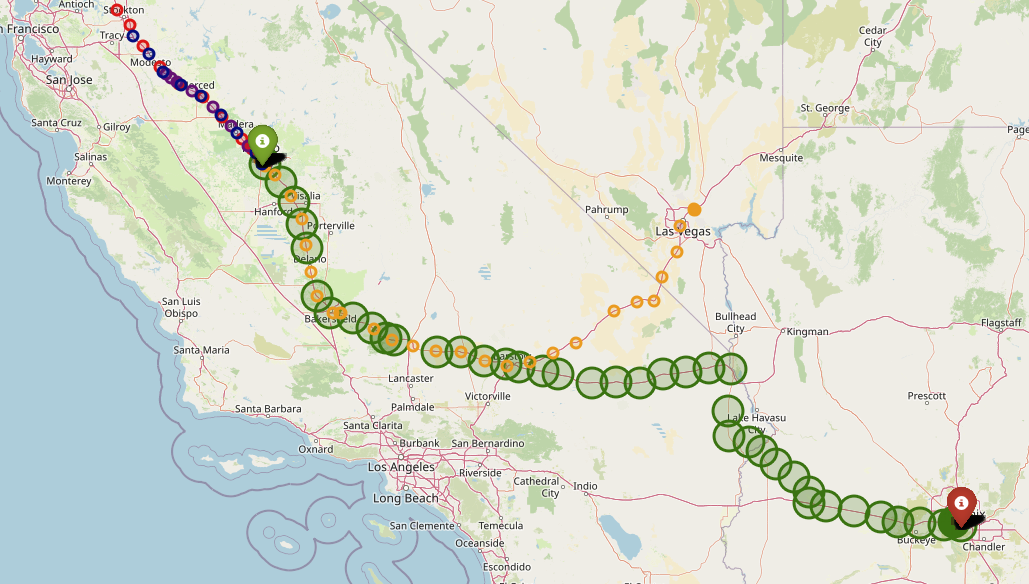}
        \caption{Correct truck selected despite opposing and deviating candidates.}
        \label{fig:multi_truck_success_1}
    \end{subfigure}%
    \hfill
    \begin{subfigure}[b]{0.32\textwidth}
        \centering
        \includegraphics[width=\textwidth]{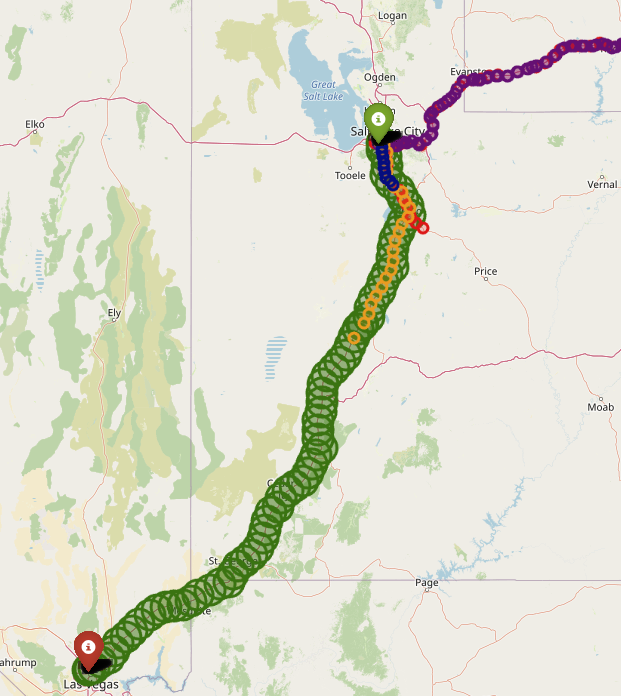}
        \caption{Correct selection when trucks initially share direction.}
        \label{fig:multi_truck_success_2}
    \end{subfigure}
    \caption{Choosing the right truck (green) among multiple candidates going to different destinations.}
    \label{fig:multi_truck_cases}
\end{figure}

\begin{figure}[!htbp]
    \centering
    \includegraphics[width=0.3\textwidth]{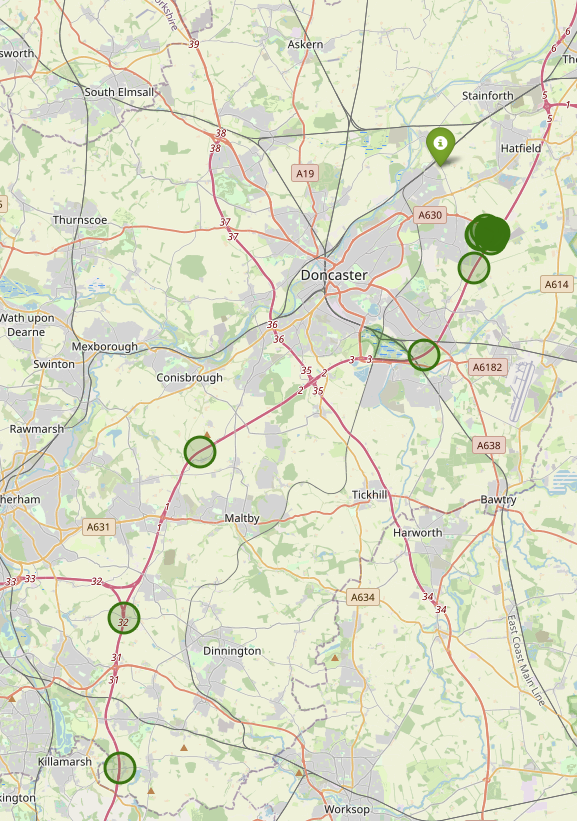}
    \caption{Robustness to geocoding errors: actual pickup is $\sim$600m from geocoded location. Model still assigns correctly using lane-based features.}
    \label{fig:robustness_to_geocode}
    \label{fig:geo_error}
\end{figure}

Fig.~\ref{fig:multi_truck_success_1} and Fig.~\ref{fig:multi_truck_success_2} show scenarios with multiple candidate trucks departing from the same origin. The model correctly identifies the right truck even when candidates initially share direction before diverging. Fig.~\ref{fig:robustness_to_geocode} demonstrates robustness to geocoding errors—the geocoded location is more than 600 meters from the actual ping cluster, yet the model still assigns correctly by relying on lane-based geospatial features.

\subsubsection{Matching variability over time}
\label{sec:match_time}

\begin{figure}[!htbp]
    \centering
    \includegraphics[width=0.4\textwidth]{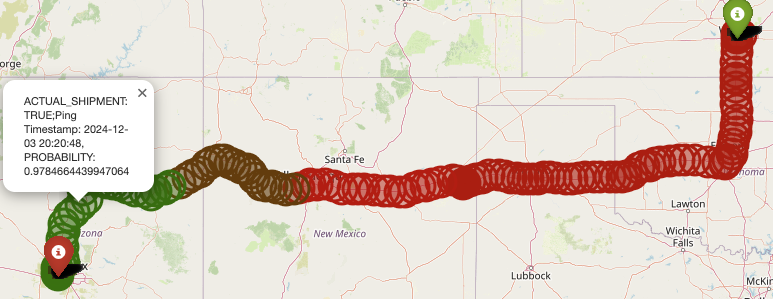}
    \caption{Probability score increases as the truck nears the destination (red: $<$0.4, green: $>$0.8).}
    \label{fig:prob-score}
\end{figure}

Beyond DPP, the number of pings received plays an important role in matching confidence. Fig.~\ref{fig:prob-score} shows that probability scores increase as trucks approach destinations, reflecting growing route similarity information. This creates a tradeoff between earlier assignment (fewer pings) and accuracy. We address this by setting minimum ping thresholds based on lane-by-lane performance observed during training.

\section{Discussion}
\label{sec:discussion}

The performance improvements of ITM 2.0 (26 p.p. precision gain in NA, 14 p.p. in EU) can be attributed to three factors validated through our ablation study. \textbf{Hexagonal spatial discretization is the critical enabler}—removing hexcell features causes precision to drop to 6-7\%, validating ping2hex as the primary driver of accuracy. The hexagonal grid's uniform neighbor relationships enable effective trajectory comparison even when trucks take slightly different paths; without spatial representation, the model resorts to temporal proximity matching, yielding false positives from trucks departing simultaneously toward different destinations. \textbf{Post-processing provides essential precision-coverage tradeoff control}—removing it drops precision by 22-35 p.p., as threshold-based filtering (Algorithm~\ref{alg:postprocessing}) prevents low-confidence assignments. \textbf{LightGBM's advantages extend beyond accuracy}—its 3× faster training and lower memory footprint are crucial for production deployment with regular retraining.

\subsection{Regional Performance Disparities}

North America achieves higher precision (55\%) than the European Union (32\%), despite both showing improvements over baselines. Three factors contribute: \textbf{training data volume bias} (more NA shipments provide richer representation), \textbf{geographic and regulatory differences} (EU has denser road networks, shorter hauls, stricter driving regulations increasing trajectory variability), and \textbf{carrier behavior heterogeneity} (standardized NA interstate trucking versus diverse EU cross-border operations). We are exploring region-specific models to address this disparity.

\subsection{Limitations and Threats to Validity}

\textbf{Short-haul limitation.} ITM 2.0 achieves ~20\% precision for shipments $\leq$ 40 km due to high spatial overlap and limited pings in metropolitan areas. Solutions include higher-resolution hexcells (H3 8-9), facility-specific features (gate IDs), or specialized methods.

\textbf{Lane data dependency.} New lanes lack historical hexcell mappings. We are exploring OSM API \cite{osrm} to synthesize lane data, though OSM routes may differ from actual trajectories.

\textbf{Evaluation constraints.} Shadow testing provides external validity but precludes randomized experiments. Performance may be influenced by temporal factors.

\textbf{Generalizability.} Results reflect our environment (full truckload, NA/EU). Performance may differ in other contexts (less-than-truckload, different regions, different ELD quality).

\subsection{Practical Implications}

By correctly matching trucks to shipments despite missing identifiers, ITM 2.0 enables hundreds more daily shipments to gain real-time tracking, improves ETA accuracy using actual truck trajectories, and supports proactive exception management through early detection of delays or off-route trucks. Methodologically, our results show that spatial discretization is more effective than raw GPS coordinates for trajectory comparison, and that hybrid ML-rule systems outperform purely algorithmic or purely rule-based approaches.

\section{Conclusion}

This paper presents Intelligent Truck Matching (ITM) 2.0, a machine learning system that addresses a critical gap in full truckload supply chain visibility: matching trucks to shipments when traditional identifiers fail. Our ping2hex spatial discretization using Uber H3 hexagonal indexing, combined with LightGBM probabilistic ranking and threshold-based post-processing, achieves 26 percentage point precision gain in North America and 14 points in Europe while doubling coverage. Deployed at Project44, the system handles medium and long-haul shipments with robustness to geocoding errors and ambiguous trajectories. Beyond the engineering contribution, our work establishes that hexagonal spatial discretization of GPS trajectories is a viable and effective representation for freight matching under uncertainty—a finding applicable to other logistics problems involving noisy geospatial data.

Future work extends in several directions. First, improving short-haul matching through higher-resolution hexcells (H3 8-9) and facility-specific features such as gate or dock identifiers. Second, exploring deep learning approaches for trajectory encoding—sequence models such as LSTMs or Transformers could capture temporal dynamics in ping sequences that fixed-feature approaches miss. Third, investigating transfer learning across regions to address the NA-EU performance disparity without requiring region-specific models. Finally, extending the framework beyond full truckload to less-than-truckload and intermodal freight, where multi-stop trajectories and shared capacity introduce additional matching complexity.

\section*{Acknowledgements}

The authors would like to thank Project44 for supporting this research and providing access to real-world data and infrastructure.

\bibliographystyle{elsarticle-num}

\end{document}